\documentclass[10pt,a4paper,twoside]{tau-book}
\usepackage[english]{babel}
\usepackage{kappa}
\usepackage{doi}

\title{EmissionNet: Air Quality Pollution Forecasting for Agriculture}

\author {Prady Saligram, Tanvir Bhathal}

\affil{Stanford University}

\institution{Stanford University}
\ftitle{Prady Saligram, Tanvir Bhathal}
\theday{March 2025} % \today
\etal{et al.}

\theabstract{Air pollution from agricultural emissions is a significant yet often overlooked contributor to environmental and public health challenges. Traditional air quality forecasting models rely on physics-based approaches, which struggle to capture complex, nonlinear pollutant interactions. In this work, we explore forecasting N$_2$O agricultural emissions through evaluating popular architectures, and proposing two novel deep learning architectures, EmissionNet (ENV) and EmissionNet-Transformer (ENT). These models leverage convolutional and transformer-based architectures to extract spatial-temporal dependencies from high-resolution emissions data.}

\begin{document}
\renewcommand{\thefigure}{\arabic{figure}}
\setcounter{figure}{0}
	
    \maketitle

    \begin{figure}[h]
        \centering
        \includegraphics[width=0.5\textwidth]{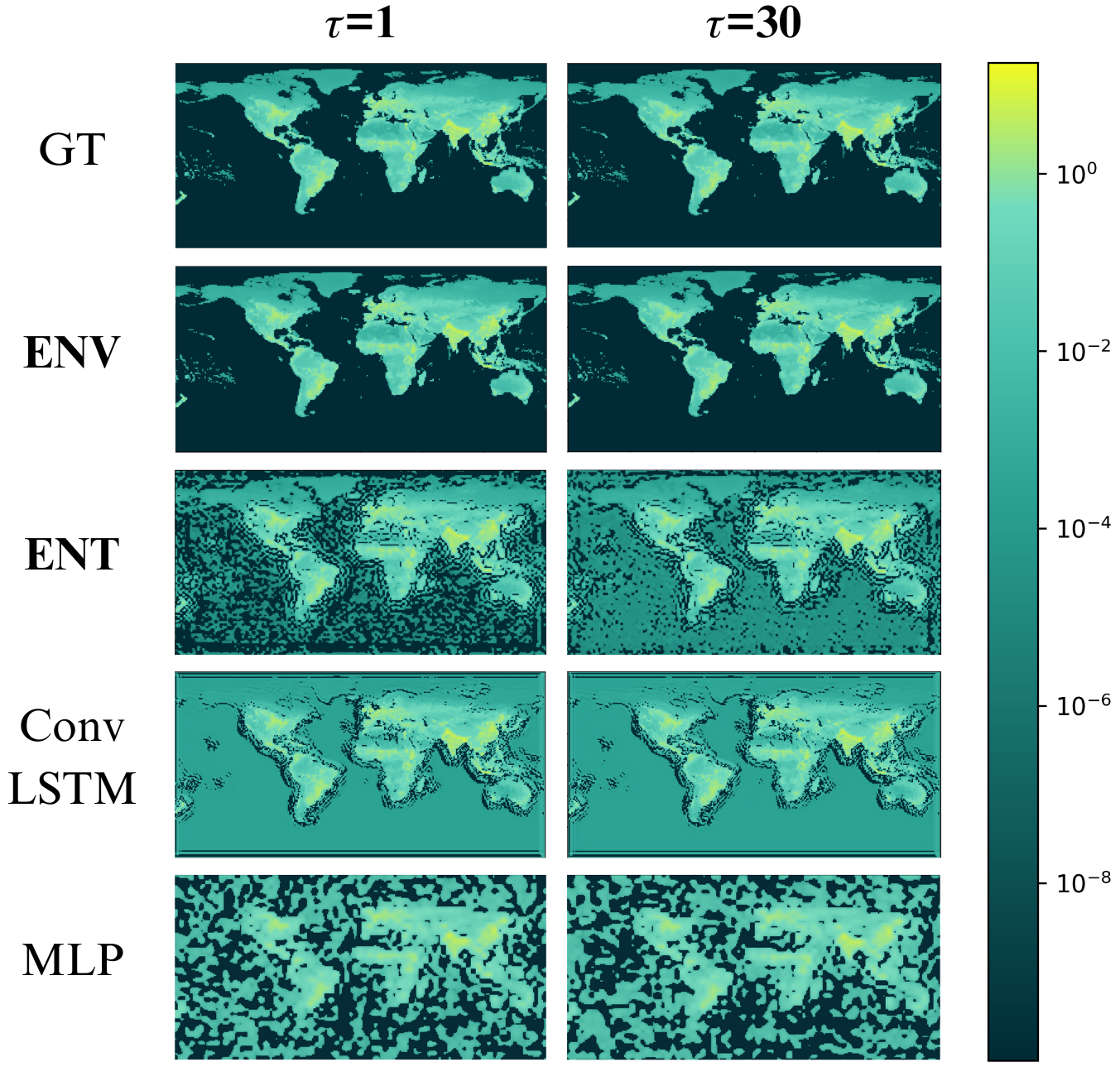}
        \caption{Baseline and Proposed Architecture Results: each unit increase of $\tau$ is equal to one month. The first row contains ground truth labels, the second/third rows display the predictions of our ENV/ENT models inputted with with 24 context frames, while the final two rows contain the predictive outputs of the baselines MLP/ConvLSTM models}
        \label{fig:main}
    \end{figure}
    
    \abscontent
    % \tableofcontents
    % Uncomment to activate the table of contents
    \thispagestyle{firststyle}

%------------------------------------------------------
% The document begins
%------------------------------------------------------
Code available on Github: \url{https://github.com/mrTSB/CS229-Final-Project}.
\vspace{-10pt}
\section{Introduction}

        \taustart{A}ir pollution stands as one of the most pressing environmental challenges, impacting human health, ecosystems, and climate stability globally. The Great Smog of London in 1952, caused by extensive coal burning, led to thousands of deaths and spurred modern air quality regulations \cite{turn0search19}. Notably, in the United States, the Clean Air Act of 1970 marked a major turning point, setting limits on pollutants such as carbon monoxide and sulfur dioxide \cite{turn0search2}. Meanwhile, in the present day, cities like Beijing, Delhi, and Los Angeles have faced ongoing air quality crises due to industrial emissions, vehicle pollution, and natural disasters like wildfires \cite{turn0news38}. Yet, amidst these high-profile urban and industrial cases, agricultural emissions—a substantial yet frequently overlooked contributor to air pollution—demand increased attention.
        
        Agricultural air pollution encompasses various pollutants, primarily, nitrous oxide (N$_{2}$O), particulate matter (PM$_{2.5}$), methane (CH$_{4}$), and carbon dioxide (CO$_{2}$), resulting from fertilizer application, livestock operations, and biomass burning. These pollutants contribute not only to respiratory and cardiovascular illnesses but also significantly influence climate change dynamics and ecosystem health \cite{turn0news17}.

        Traditional air quality forecasting methods, such as the EPA's Community Multiscale Air Quality Model (CMAQ), AERMOD, and NASA's GEOS-CF system, have made significant strides by modeling atmospheric chemistry through differential equations and transport phenomena \cite{epa_scram} \cite{epa_cmaq} \cite{nasa_geoscf}. However, recent advances in deep learning have demonstrated the potential to surpass these conventional frameworks. Studies have introduced deep learning systems capable of capturing the nonlinear response of air quality to emissions changes \cite{xing2020}, while models such as FuXi Weather~\cite{fuxi} and Aurora~\cite{aurora} have demonstrated state-of-the-art performance in predicting atmospheric dynamics. Despite their effectiveness, these models remain limited by deterministic outputs and recurrent architectures that fail to fully address the stochastic nature of pollutant dispersion and accumulation.

        Our research focuses on forecasting N$_2$O emissions, a major air pollutant with significant impacts on air quality. Using historical global emissions data \cite{Tian2020}, we employ a rolling-window approach, leveraging 24 months of prior observations from five key emissions—CH$_4$, CO$_2$, N$_2$O, CO$_2$bio, and GWA—to predict N$_2$O emissions for the following month. Formally, given a sequence of monthly observations  
        \[
        X = \{x_t, x_{t+1}, \dots, x_{t+23}\},
        \]  
        our model predicts the
        \[
        x_{t+24},
        \]  
        N$_2$O emissions for the next month.

        Our approach integrates high-resolution emissions data from the EDGAR database with machine learning techniques to develop a spatiotemporal prediction framework \cite{edgar}. This project makes two key contributions: first, evaluating the effectiveness of baseline models such as MLPs and ConvLSTMs; second, proposing hybrid architectures that combine ConvLSTM, DenseNet layers, and more. Our novel design EmissionNet (ENV) captures both spatial and temporal patterns in agricultural emissions, improving forecasting accuracy beyond traditional physics-based models by reducing our Test MSE from 0.001567 to 0.0006836. This research aims to enhance N$_2$O and pollutant predictions, offering valuable insights for environmental policy and agricultural management.

        % a deep learning-based spatiotemporal forecasting model designed to overcome the limitations of traditional air quality prediction systems. By leveraging high-dimensional tensor representations of emission data from the EDGAR database—specifically targeting agricultural emissions—and integrating them with time-series air quality index (AQI) data from Kaggle, we aim to build a predictive framework capable of capturing complex pollution dynamics across both space and time. At the core of our architecture lies a hybrid combination of ConvLSTM and DenseNet layers, designed to simultaneously model long-term temporal dependencies and preserve detailed spatial features across layers. Further, implicit deep supervision modules with residual connections facilitate gradient flow across layers, allowing the network to learn refined feature representations over time. A channel attention mechanism is integrated to dynamically recalibrate feature importance using global average pooling and fully connected layers, ensuring that the model focuses on the most relevant emissions and spatial regions for precise air quality forecasting.
\section{Related Work}

We briefly analyze current state-of-the-art approaches to related topics, such as weather and general pollution forecasting. While we did not find any specific work on models for predicting agricultural pollution, the concepts and techniques from pollution and weather forecasting are highly similar.

Google's DeepMind introduced \textit{GenCast}, an AI-driven model that predicts weather conditions up to 15 days in advance, outperforming traditional forecasting systems in both accuracy and speed. It utilizes diffusion models to address physical inconsistencies \cite{Price2024}.

\textit{FuXi Weather} takes a recursive approach, using each predicted latent representation to generate the next state. However, this method suffers from significant compounding errors \cite{fuxi}.

\textit{Aurora}, a 1-billion-parameter foundation model developed by Microsoft for weather modeling, has over a million hours of training on diverse datasets. It employs a similar architecture to \textit{FuXi} but enhances accuracy by using encoders and decoders to better capture state dynamics \cite{aurora}.

Not much is known about the European Centre for Medium-Range Weather Forecasts' system, which claims to achieve a 20\% improvement in forecast accuracy. However, available information suggests they employ an Ensemble Prediction System approach \cite{IntegratedForecastSystemWiki}.

Furthermore, studies indicate that for predicting agricultural groundwater, Multi-Layer Perceptron (MLP) models perform best. This conclusion was drawn from comparisons with various models, including Random Forest, Gaussian Process Regression, Random Subspace, and MLP models \cite{water}. Additionally, multiple studies have concluded that boosting techniques, such as XGBoost \cite{xgboost}, show significant promise in predicting air quality and atmospheric molecule concentrations \cite{Li2023, RAVINDIRAN2023139518, turn0search0}.

\section{Dataset and Features}

We used emissions data from the EU’s Emissions Database for Global Atmospheric Research (EDGAR) \cite{edgar}, which provides high-resolution global GHG emissions from 2000 to 2023. The dataset includes \texttt{netCDF} files with monthly grid maps (e.g., emissions in Mg/month, fluxes in kg/m$^2$/s) across various sectors.

Each annual \texttt{netCDF} file (72 GB) contains 12 monthly slices. The spatial grid spans latitudes $[-90^\circ, +90^\circ]$ (\texttt{[-900, 900]}) and longitudes $[-180^\circ, +180^\circ]$ (\texttt{[-1800, 1800]}) at a $0.1^\circ$ resolution. The dataset covers emissions and fluxes for five key molecules (CH$_4$, CO$_2$, N$_2$O, CO$_2$bio, GWA) and includes total national emissions in kilotons.

% Given the extensive data size (72 GB), storage and processing posed significant computational challenges. To prevent memory overload, we employed memory-mapped files (\texttt{numpy.memmap}), which enabled efficient data handling. This approach provided two key advantages:

% \begin{itemize}
%     \item \textbf{Optimized Read/Write Operations:} By reducing frequent data transfers between memory and disk, this method minimized processing overhead.
    
%     \item \textbf{Memory Management and Data Integrity:} Each monthly data chunk was explicitly flushed to disk after being written, ensuring data integrity and preventing memory overflow. \textit{This technique was particularly crucial during model training, as even large-scale computing clusters frequently encountered memory allocation issues}.
% \end{itemize}

% By leveraging a memory-mapped architecture (\texttt{numpy.memmap}), we efficiently processed the high-resolution dataset (72 GB) while preserving spatial and temporal fidelity. Memory-mapped files allowed data flushing after writing, preventing memory overflow—an essential solution for handling large models and mitigating memory constraints during training.

Using \texttt{xarray}, \texttt{netCDF4}, and \texttt{numpy}, we concatenated emissions and flux data over time, forming a five-dimensional tensor \cite{mckinney2010data}. A direct multi-layer neural network on this tensor would yield an impractically large model ($\sim$480M parameters). To reduce computational costs, we applied $0.3^\circ \times 0.3^\circ$ spatial pooling and discarded the flux dimension, as predicting flux values was deemed not relevant. This preprocessing step reduced the final tensor to:
\[
\text{Shape: } (5, 288, 600, 1200),
\]\\
\\where:
\begin{itemize}
    \item 5 represents molecular emissions (CH$_4$, CO$_2$, N$_2$O, CO$_2$bio, GWA),
    \item 288 corresponds to 24 years $\times$ 12 months,
    \item 600 $\times$ 1200 defines the latitude-longitude grid.
\end{itemize}
   % \textbf{Statistical Validation and Autocorrelation Analysis:}  
   %  To evaluate the suitability of the dataset for time-series forecasting, we conducted statistical validation and autocorrelation analysis:
   %  \begin{itemize}
   %      \item \textbf{Autocorrelation Analysis:} Strong seasonal periodicity was observed, corresponding to agricultural cycles such as livestock patterns and crop seasons, with high autocorrelation coefficients across the dataset.
   %      \item \textbf{Cross-Validation:} Aggregated emissions from the processed dataset were compared against sector-level totals published by EDGAR, confirming data consistency and accuracy.
   %  \end{itemize}
   %  The presence of robust cyclical patterns in agricultural emissions supports its use for spatiotemporal forecasting with deep learning models.

    % \textbf{Dimensionality Reduction:}  
    % A direct implementation of a multi-layer neural network on the original tensor would result in an impractically large model with approximately 480 million parameters. To mitigate computational constraints, we applied spatial pooling, aggregating emissions over $3^\circ \times 3^\circ$ latitude-longitude grid cells. Additionally, we determined that predicting flux values was less critical than emissions, allowing us to discard the flux dimension. This preprocessing step reduced the final tensor to:
    % \[
    % \text{Shape: } (5,\,288,\,600,\,1200),
    % \]
    % significantly improving computational efficiency while preserving essential spatiotemporal patterns.
 To cross validate, aggregated emissions from the processed dataset were compared against sector-level totals published by EDGAR, confirming data consistency and accuracy.
\noindent The final dataset was partitioned into three subsets using a 90-10-10 split:

\begin{itemize}
    \item \textbf{Training Set:} January 2000 to March 2019 (231 months)
    \item \textbf{Validation Set:} April 2019 to July 2021 (28 months)
    \item \textbf{Test Set:} August 2021 to January 2024 (29 months)
\end{itemize}

Below in Figure \ref{fig:n2o} we map the Global N$_2$O emissions in 2010. More data visualizations are provided in Figures Figures~\ref{fig:co2}-\ref{fig:gwp}.

\begin{figure}[h]
    \centering
    \includegraphics[width=0.5\textwidth]{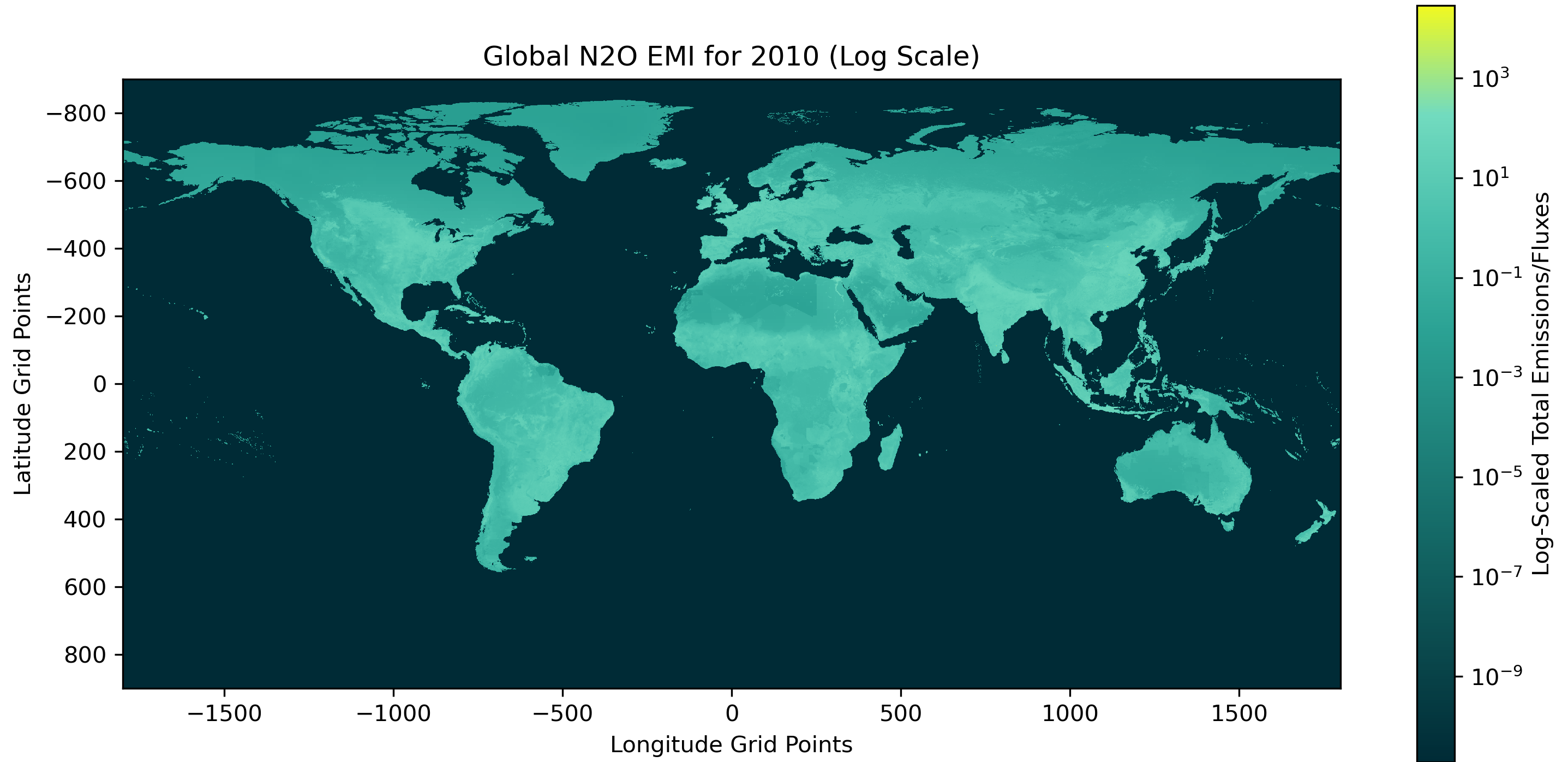}
    \caption{2010 Global N$_2$O Emissions}
    \label{fig:n2o}
\end{figure}

\vspace{-15pt}
\section{Methods}

\subsection{Preliminaries}
To effectively benchmark our proposed method, we establish robust baseline models that range from simple to progressively more sophisticated architectures. These models sequentially escalate in complexity, enabling us to analyze incremental improvements and clearly justify architectural enhancements. We first explore related technologies and techniques to build our baselines on.

Recurrent Neural Networks (RNNs), including Long Short-Term Memory (LSTM) and Gated Recurrent Units (GRU) \cite{cho2014learning}, are commonly used for time-series forecasting due to their ability to capture temporal dependencies. However, they require extensive historical sequences, leading to high computational costs \cite{hochreiter1997long, chung2014empirical}. Conversely, Convolutional Neural Networks (CNNs) excel in spatial feature extraction but lack explicit temporal modeling, making them insufficient for spatiotemporal forecasting.

% Channel attention mechanisms \cite{hu2018squeeze} refine feature importance by computing global average pooling across feature maps:
% \begin{equation}
% z_c = \frac{1}{H \times W}\sum_{i=1}^{H}\sum_{j=1}^{W}u_c(i,j),
% \end{equation}
% followed by a gating mechanism:
% \begin{equation}
% s_c = \sigma(\mathbf{W}_2\delta(\mathbf{W}_1 \mathbf{z}_c)),
% \end{equation}
% where $\delta$ is ReLU, $\sigma$ is sigmoid activation, and $\mathbf{W}_1$, $\mathbf{W}_2$ are learned parameters. The recalibrated feature map is:
% \begin{equation}
% \mathbf{X}_{\text{out}} = \mathbf{s}_c \circ \mathbf{u}_c.
% \end{equation}

\textbf{MLP Baselines.} The MLP baselines leveraged DenseNet \cite{huang2017densely} for spatial feature extraction, using dense connections to mitigate the vanishing gradient problem and reduce parameter redundancy. Each layer receives feature maps from all preceding layers:

\begin{equation}
X_l = H_l(\text{concat}[X_{l-1}, X_{l-2}, \dots, X_0]),
\end{equation}

where $H_l(\cdot)$ includes batch normalization, ReLU activation, and convolutions.

Given MLPs' success in forecasting agricultural water quality \cite{water}, we implemented an initial MLP with two hidden layers (256 neurons, ReLU). We extended this to a deeper MLP (512, 512, 128 neurons) to better capture spatiotemporal dependencies in $24$-month × spatial × $5$-channel inputs. The added capacity was hypothesized to improve accuracy by learning complex nonlinear patterns.

% \textbf{LSTM Baselines.} To explicitly account for temporal relationships in the data, we initially explored an LSTM architecture consisting of a single-layer LSTM with 128 hidden units, followed by a fully-connected output layer mapping the hidden state to the spatial prediction grid. To address observed limitations in capturing longer-range dependencies and the complex nature of temporal patterns, we subsequently implemented a second, deeper LSTM baseline. This second model included two stacked LSTM layers, each containing 256 hidden units. The expanded temporal depth (multi-layer LSTM) and increased hidden dimensionality were hypothesized to better capture more nuanced temporal correlations present within the emissions data.

\textbf{ConvLSTM Baselines.} ConvLSTM \cite{shi2015convolutional} integrates convolution into LSTM cells to capture spatial-temporal dependencies. Each ConvLSTM cell applies convolutions in both input and recurrent transformations:

\begin{align}
\mathbf{i}_t &= \sigma(\mathbf{W}_{xi} * \mathbf{X}_t + \mathbf{W}_{hi} * \mathbf{H}_{t-1} + \mathbf{W}_{ci} \circ \mathbf{C}_{t-1} + \mathbf{b}_i), \\
\mathbf{f}_t &= \sigma(\mathbf{W}_{xf} * \mathbf{X}_t + \mathbf{W}_{hf} * \mathbf{H}_{t-1} + \mathbf{W}_{cf} \circ \mathbf{C}_{t-1} + \mathbf{b}_f), \\
\mathbf{C}_t &= \mathbf{f}_t \circ \mathbf{C}_{t-1} + \mathbf{i}_t \circ \tanh(\mathbf{W}_{xc} * \mathbf{X}_t + \mathbf{W}_{hc} * \mathbf{H}_{t-1} + \mathbf{b}_c), \\
\mathbf{o}_t &= \sigma(\mathbf{W}_{xo} * \mathbf{X}_t + \mathbf{W}_{ho} * \mathbf{H}_{t-1} + \mathbf{W}_{co} \circ \mathbf{C}_t + \mathbf{b}_o), \\
\mathbf{H}_t &= \mathbf{o}_t \circ \tanh(\mathbf{C}_t),
\end{align}

where $*$ denotes convolution, $\circ$ element-wise multiplication, and $\sigma$ the sigmoid function. We employed ConvLSTM models with progressively increasing complexity to identify optimal capacity and receptive fields:
\begin{enumerate}
    \item We first implemented a single ConvLSTM layer with 8 hidden channels and a $3\times3$ convolution kernel, offering a foundational model to capture spatial-temporal correlations with minimal complexity.
    \item To enhance feature extraction, we introduced a second ConvLSTM layer with 16 hidden channels per layer. Our hypothesis was that deeper stacking would improve spatial-temporal learning, particularly for complex emission patterns, by capturing hierarchical representations across time.
    \item We further expanded the model to three stacked ConvLSTM layers with 32 hidden channels per layer and larger $5\times5$ kernels. We hypothesized that larger receptive fields and increased capacity would enable better long-range spatial dependency modeling and stronger generalization for emissions forecasting.
\end{enumerate}

Each successive baseline model iteration served the explicit purpose of evaluating whether increased complexity—in terms of depth, hidden dimensionality, and receptive field—would significantly enhance predictive performance on N$_2$O emission forecasting, thus setting the stage for our proposed advanced methodologies.

\subsection{Proposed Architecture: EmissionNet}
We propose two novel architectures that build upon state-of-the-art (SOTA) techniques, addressing limitations in baseline models to optimize forecasting performance. \cite{tensorflow2015-whitepaper} \cite{paszke2019pytorch}\\

\begin{figure}[h]
    \centering
    \includegraphics[width=0.5\textwidth]{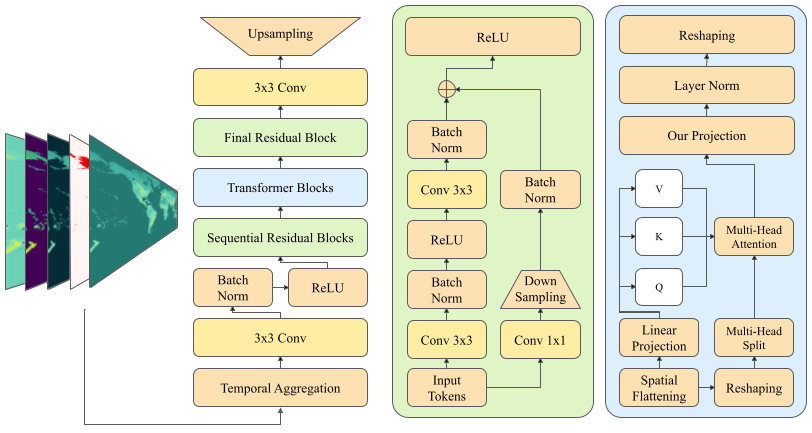}
    \caption{ENT Architecture}
    \label{fig:entv4}
\end{figure}

First, we propose \textbf{EmissionNet-Transformer (ENT)}, an enhanced spatiotemporal forecasting model integrating convolutional residual blocks with a transformer-based attention mechanism (Figure \ref{fig:entv4}). This architecture is designed to capture both local spatial features and long-range dependencies for improved N$_2$O emission prediction.

\paragraph{Initial Convolution.} 
The input sequence, consisting of five molecular emission channels, is first processed through a $3\times3$ convolutional layer with 64 filters, followed by batch normalization and ReLU activation to extract low-level spatial features.

\paragraph{Residual Blocks.} 
Three residual blocks progressively refine spatial representations. The first residual block applies a $3\times3$ convolution with stride 2 to downsample the feature map, followed by another $3\times3$ convolution for feature extraction. A $1\times1$ convolutional downsampling layer ensures identity mapping in the shortcut connection. The second residual block maintains spatial resolution while deepening feature extraction. The third residual block reconstructs the feature map, reducing the number of channels back to 64.

\paragraph{Transformer-Based Attention.} 
A lightweight \textit{FlashTransformerDecoder} module refines feature representations by applying self-attention to spatial embeddings. The attention mechanism includes linear projections for queries, keys, and values, followed by dropout and layer normalization. This module enables the model to capture long-range dependencies between emissions, improving forecasting accuracy.

\paragraph{Final Convolution and Upsampling.} 
The processed features pass through a $1\times1$ convolutional layer to generate the final N$_2$O emission predictions. A bilinear upsampling layer restores the spatial resolution to $(150, 300)$, ensuring fine-grained prediction accuracy.

This hybrid convolutional-transformer architecture effectively combines spatial feature extraction with attention-based global modeling, improving emissions forecasting beyond conventional deep learning models.\\

\begin{figure}[h]
    \centering
    \includegraphics[width=0.5\textwidth]{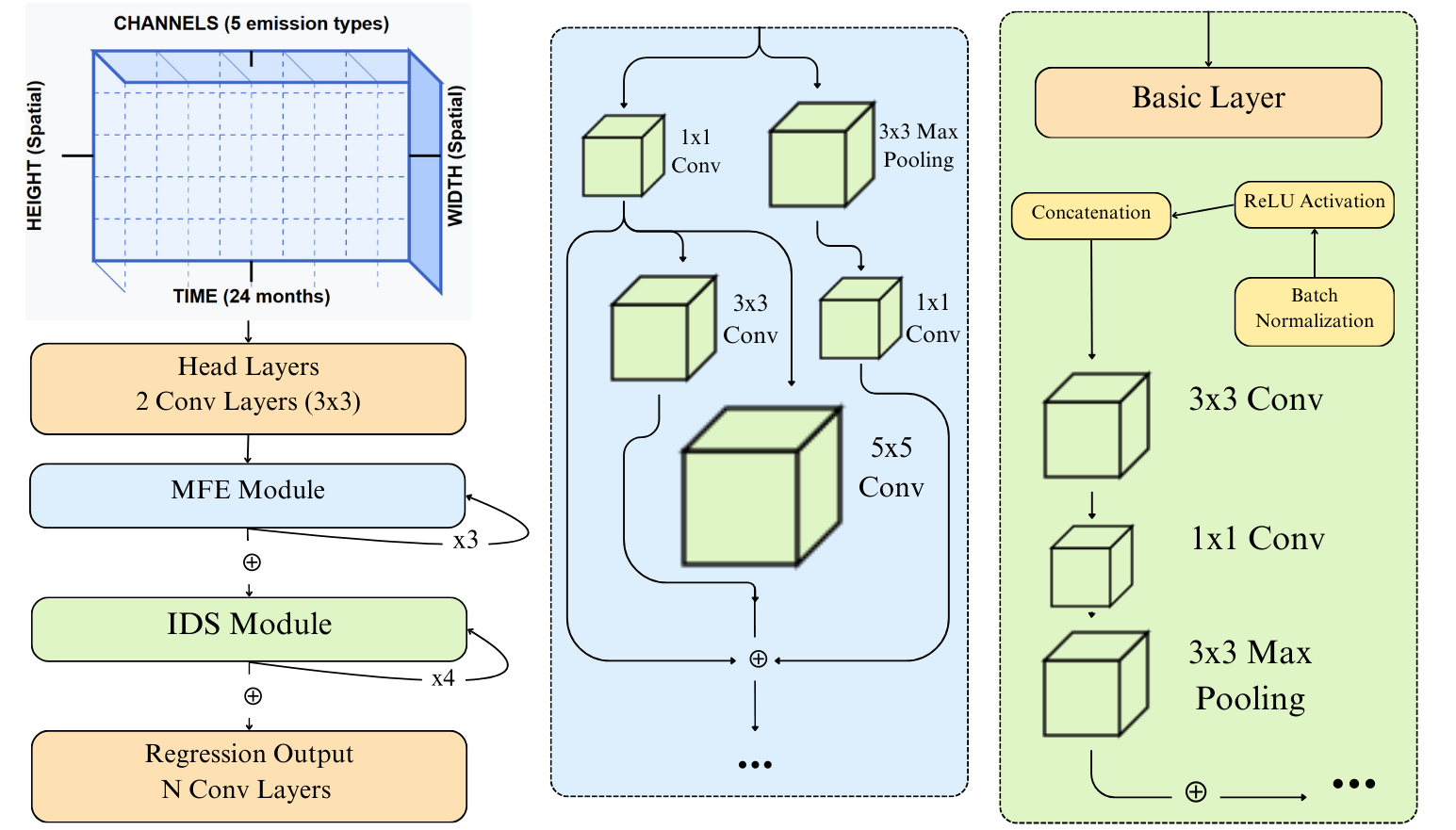}
    \caption{ENV Architecture}
    \label{fig:env1}
\end{figure}

We propose \textbf{EmissionNet (ENV)}, a convolutional architecture explicitly tailored for regression tasks involving spatial-temporal emission prediction (Figure \ref{fig:env1}). Inspired by multi-scale feature extraction and implicit deep supervision principles, EmissionNet effectively captures intricate spatial patterns and temporal dependencies inherent in emissions data \cite{Li}.

The architecture consists of three primary components: two convolutional head layers, multi-scale feature extraction (MFE) modules, and implicit deep supervision modules enhanced by channel attention mechanisms.

\paragraph{Head Layers.} 
Initially, input sequences (e.g., 24-month windows with five molecular emission channels) are processed through two convolutional head layers. To preserve spatial details, these layers utilize 3×3 convolution filters with stride 2, replacing conventional pooling operations.

\paragraph{Multi-scale Feature Extraction.}
The processed features are then expanded by three MFE modules. Each MFE module captures rich spatial details at varying receptive fields through parallel convolutional branches:
\begin{equation}
    B = [H_{1\times1}(A),\, H_{3\times3}(A),\, H_{5\times5}(A),\, H_{\text{pooling}}(A)],
\end{equation}
where $A$ represents the input feature map, and $[\,\cdot\,]$ denotes concatenation of feature maps. The branches utilize convolutions of kernel sizes 1×1 (for dimensionality reduction), 3×3, and 5×5, along with a parallel 3×3 max pooling operation.

\paragraph{Implicit Deep Supervision.}
Next, features flow through four implicit deep supervision modules, each composed of sequential \textit{basic layers}. Each basic layer integrates all previous layers' outputs via concatenation, as formulated below:
\begin{equation}
    x_l = H_l([x_{l-1}, x_{l-2}, \dots, x_0]),
\end{equation}
where $H_l(\cdot)$ represents a batch normalization (BN) operation, followed by a ReLU activation and 3×3 convolution. This densely connected structure improves gradient flow and enhances feature representations across the network. To prevent dimensional explosion, each module concludes with a 1×1 convolution followed by a 2×2 max pooling operation with stride 2 for dimension reduction.

\paragraph{Channel Attention Mechanism.}
To further refine feature relevance, we integrate a channel attention mechanism between consecutive basic layers within each implicit deep supervision module. This mechanism recalibrates feature maps by performing global average pooling followed by channel-specific recalibration weights computed via two fully connected layers:
\begin{align}
    z_c &= G(u_c) = \frac{1}{H\times W}\sum_{i=0}^{H-1}\sum_{j=0}^{W-1}u_c(i,j), \\
    s_c &= \sigma(w_2 \otimes \delta(w_1 \otimes z_c)), \\
    x_{out} &= s_c \cdot u_c,
\end{align}
where $u_c(i,j)$ is the feature value at spatial position $(i,j)$ of channel $c$, $G(\cdot)$ denotes global average pooling, $\sigma(\cdot)$ and $\delta(\cdot)$ are sigmoid and ReLU activation functions, respectively, and $\otimes$ indicates convolution operations. 

\paragraph{Regression Output Layer.}
Finally, the refined feature maps pass through a series of convolutional layers without classification heads, concluding with a single convolutional layer that outputs spatially resolved predictions of the N$_2$O emission values for the next month. This configuration is optimized explicitly for high-fidelity regression, aligning precisely with the spatiotemporal prediction requirements of our emissions dataset.

\section{Results and Discussion}
To initially validate the use of ML and DL techniques, we evaluate the suitability of the dataset for time-series forecasting through conducting statistical validation and autocorrelation analysis. We found strong seasonal periodicity, corresponding to agricultural cycles such as livestock patterns and crop seasons, with high autocorrelation coefficients across the dataset.

The presence of robust cyclical patterns in agricultural emissions supports its use for spatiotemporal forecasting with deep learning models.

The primary metric we measure is the mean square error (MSE) of both proposed architectures, ENT and ENV, against both baselines, MLP and ConvLSTM \cite{scikit-learn}. As there were multiple variations of the MLP and ConvLSTM baselines, we are only looking at the best performing baselines. All models are trained, validated, and tested on the same aforementioned splits of the processed EDGAR dataset, with the same adaptive learning rate algorithm.

We conducted hyperparameter sweeps to optimize each architecture, identifying the optimal dynamic learning rates of $3\times10^{-3}$, with warmup ratios of 0.2–0.3 and a weight decay of $1\times10^{-4}$ (Table \ref{tab:hyperparam_mse}). Training was run for 200 epochs, beyond which improvements were negligible. To accommodate increasing model complexity, batch sizes were reduced from 32 to 8, ensuring stable training across deeper architectures.

We see that ConvLSTM are more effective than MLP with a Test MSE or 0.00156 compared to 0.06288 (Table \ref{tab:mse_comparison}). However, in comparison to this ConvLSTM approach, ENT was significantly better with a Test MSE of 0.00068, and ENV reached a Test MSE of 0.00010.

\begin{table}[h]
    \centering
    \caption{Mean Squared Error (MSE) Comparison Across Models}
    \label{tab:mse_comparison}
    \begin{tabular}{lccc}
        \hline
        \textbf{Model} & \textbf{Train MSE} & \textbf{Validation MSE} & \textbf{Test MSE} \\
        \hline
        MLP & 0.33551 & 0.38383 & 0.06288 \\
        ConvLSTM & 0.00329 & 0.00367 & 0.00156 \\
        ENT & 0.00262 & 0.00285 & 0.00068 \\
        ENV & \textbf{0.00082} & \textbf{0.00092} & \textbf{0.00010} \\
        \hline
    \end{tabular}
\end{table}

We observe this trend across both the Train and Validation sets, with ENV consistently achieving the best performance and MLP the worst (Table \ref{tab:mse_comparison}). Furthermore, when looking at the visual map representations of the data, we observe that ENV is nearly identical and is the only model which does not degrade in quality throughout predictions in the test set (Figure \ref{fig:main}). Previous studies that identified MLPs as the best models primarily compared them to simpler architectures \cite{water}. Thus, it is not unsurpirsing that when MLPs are evaluated against more advanced architectures such as ConvLSTM, ENT, and ENV, they perform worse.

% \begin{figure}[h]
%     \centering
%     \includegraphics[width=0.5\textwidth]{mse.png}
%     \caption{Train, Validation, and Test MSEs}
%     \label{fig:mse}
% \end{figure}

\begin{figure}[h]
    \centering
    \includegraphics[width=0.5\textwidth]{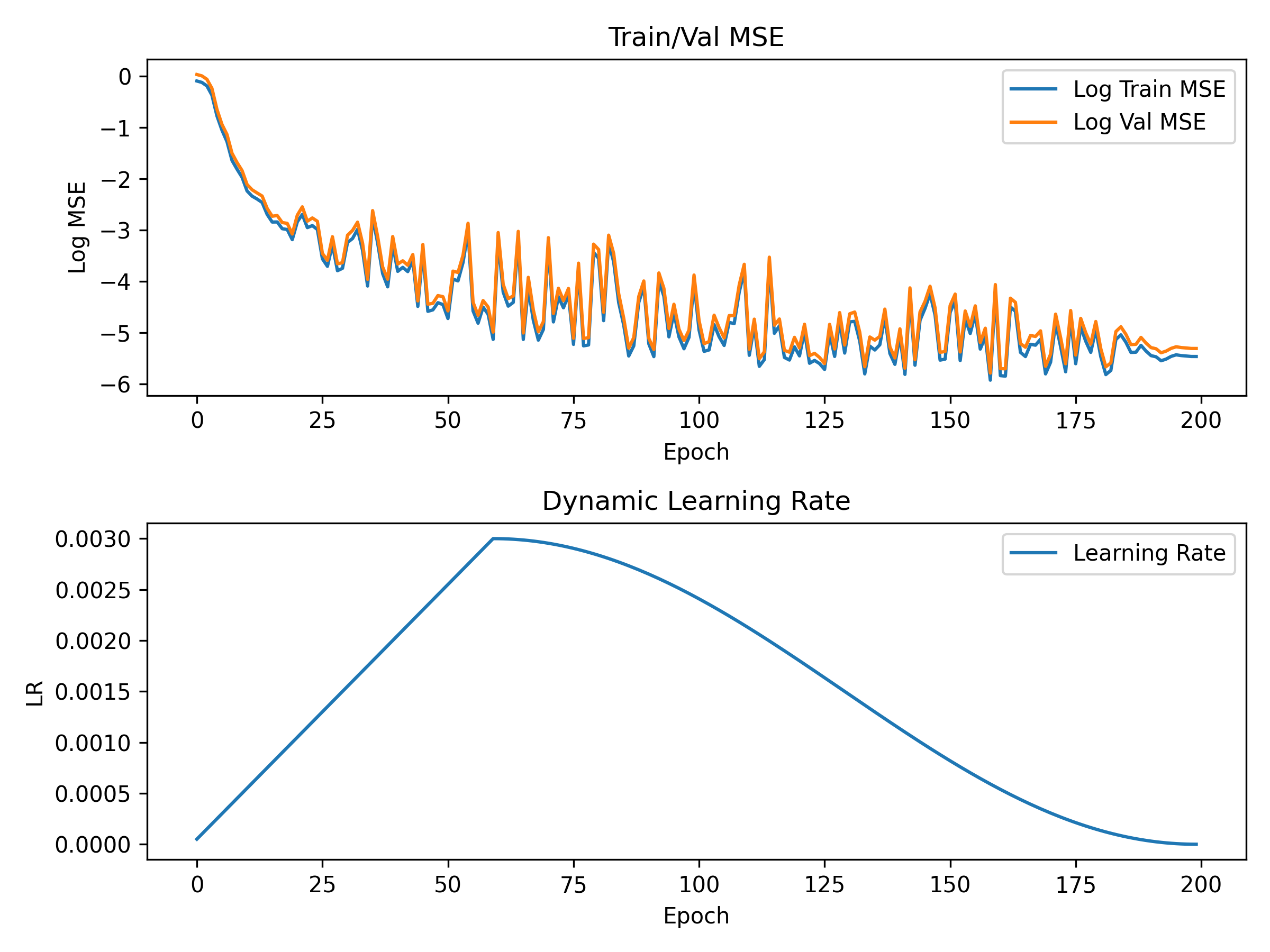}
    \caption{ENV Train and Validation MSE over Training Epochs}
    \label{fig:env1_loss}
\end{figure}

Furthermore, we observe potential reasons why the MLP and ConvLSTM architectures may perform worse than ENT and ENV. Upon examining the Training and Validation MSE of all architectures, we notice an interesting trend\textemdash MLP and ConvLSTM stop learning too early (Figures~\ref{fig:env1_loss}, \ref{fig:entv4_loss}, \ref{fig:convlstm_loss}, \ref{fig:mlp_loss}) \cite{hunter2007matplotlib}. 

Observing the logarithmically scaled MSE axis, we see that ENT and ENV begin converging after approximately 150 epochs, reaching log MSE values between -4 and -6. In contrast, ConvLSTM and MLP start converging much earlier, around 10 epochs, with log MSE values between -1.5 and -2.5. Due to the logarithmic nature of this axis, this represents a substantial difference, highlighting that a longer active learning phase significantly enhances model performance.

Additionally, from Figure~\ref{fig:main} it is clear that while ConvLSTM excels at capturing high‐gradient regions (e.g., the abrupt contrasts at continental boundaries), its predictions exhibit a near‐constant offset over oceans, where gradients are gentler. This bias suggests that, though it can effectively learn spatiotemporal patterns, it can also form overly rigid assumptions about regions lacking strong emission gradients (e.g., open water).

The more advanced ENT and ENV architectures overcome these deficiencies by combining multi‐scale feature extraction with either global attention or implicit deep supervision where ENT leverages its transformer decoder to aggregate long‐range dependencies across the spatial grid, mitigating localized biases such as the ocean offset seen in ConvLSTM. Meanwhile, ENV (with MFE modules and channel attention) achieves a more consistent emission map in low‐gradient regions by enforcing multi‐scale consistency at each stage. Overall, this spurs finer capture of both strongly differentiated continents and smoothly varying open‐water areas, as visualized in their final predictions.

\begin{figure}[h]
    \centering
    \includegraphics[width=0.5\textwidth]{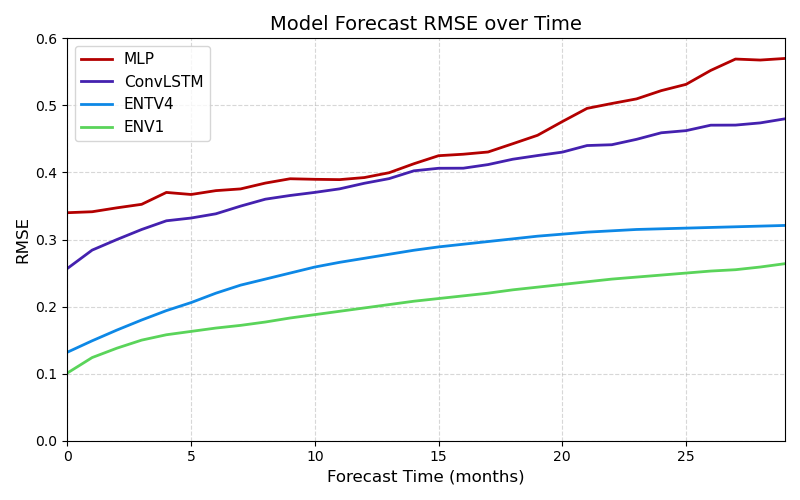}
    \caption{Auto-Regressive RMSE over Time}
    \label{fig:rmse}
\end{figure}

The compounding errors over time stem from the auto-regressive nature of the forecasting process, where each future prediction is conditioned on previous model outputs rather than ground truth values (Figure \ref{fig:rmse}). This causes errors to accumulate, leading to a growing RMSE. Despite this, ENT approaches an asymptote, while MLP, ConvLSTM, and ENV continue growing.

The faster-growing RMSE loss of ENV compared to ENT can be attributed to their architectural differences. ENT integrates a transformer-based attention mechanism, which enables it to capture long-range dependencies in emissions data, allowing for more stable long-term predictions. As a result, its RMSE plateaus over time, suggesting it is approaching an asymptotic limit in its forecasting errors. This is especially impressive given ENT is the lowest parameter model (Table \ref{tab:param_counts}). 

In contrast, ENV relies mostly on convolutional and multi-scale feature extraction modules, which are well-suited for local spatial dependencies but less effective at modeling long-term temporal relationships. This limitation leads to accumulating errors as the forecast horizon extends, causing the RMSE to keep increasing rather than stabilizing. The absence of a global attention mechanism means ENV struggles to generalize patterns over time, leading to its diverging error trend.

\section{Conclusion}
We showcase that ENV is a robust architecture for spatiotemporal agricultural emission forecasting, capable of capturing the full spectrum of emission dynamics across tabularized gridmap data. By integrating convolutional head layers with residual blocks, transformer modules, and sampling strategies, ENV accurately maps the intricate spatiotemporal functions that govern both differentiably dynamic and static emissions. Unlike conventional models that primarily excel in capturing dynamic behaviors over densely populated or continental regions, our architecture effectively generalizes to static emission sources over the ocean. 

Our evaluation reveals that ENV not only surpasses SOTA methods such as ConvLSTM and traditional sole channel attention modules, but it also demonstrates superior functional consistency over short- and medium-term forecasting horizons. The inclusion of flash attention mechanisms within the transformer module enables ENV to capture long-range dependencies and subtle spatial correlations that are essential for reliable emission mapping.

ENV sets a new benchmark for emission forecasting, providing a scalable and accurate solution that can be readily adapted to diverse geographic and temporal contexts. Future research will focus on further expanding the model’s parameter space and incorporating additional environmental variables, paving the way for even more refined and comprehensive emission prediction systems. Furthermore, with more time and compute, we could reverse pooling steps in the data, to have ultra-fine spatial resolution.
%------------------------------------------------------

\appendix
\section*{Appendix}
\addcontentsline{toc}{section}{Appendix}

\subsection*{7.1 Data Referencing}
Aurora~\cite{aurora} is pre-trained on the ERA5 dataset, which includes 4 surface variables and 13 levels of 5 atmospheric variables, covering the period from January 1, 1917, to December 31, 2020. This dataset amounts to 105.5~TB of data. Additionally, Aurora is pre-trained on various other sources of weather data (e.g., HRES-0.25 forecasts, IFS-ENS-0.25, and GFS forecasts), which collectively total 1268.12~TB. During inference, Aurora is conditioned on 4 surface variables and 5 atmospheric variables across 13 levels.

On the other hand, our dataset comprises strictly the agricultural-based emissions from the EU’s Emissions Database for Global Atmospheric Research (EDGAR)~\cite{edgar}. Our dataset contains a more limited timeframe of January 1, 2000 to December 31, 2023 amounting to a total storage of ~72 GB when storing all 5 emissions across the gridmap in a single NumPy tensor. 

The parameter counts for these models (computed in detail from our architectures) are summarized in Table~\ref{tab:param_counts}.

\begin{table}[h]
    \centering
    \caption{Parameter Counts of Current Models}
    \label{tab:param_counts}
    \begin{tabular}{lc}
        \hline
        \textbf{Model} & \textbf{Parameter Count} \\
        \hline
        ENV            & $\sim 11.7$ M \\
        ENT  & $\sim 0.71$ M \\
        MLP              & $\sim 32$ M \\
        ConvLSTM   & $\sim 1.3$ M \\
        \hline
    \end{tabular}
\end{table}

\subsection*{7.2 Hyperparameter Sweeps}
We conducted extensive hyperparameter sweeps across our baseline models. The hyperparameters of interest include the learning rate, batch size, context window size, forecast horizon, number of training epochs, warmup ratio for the dynamic learning rate, and optimizer. Table~\ref{tab:hyperparam_mse} summarizes the selected configurations for each model in our experiments, where LR is the Learning Rate, BS is the Batch Size, CW is the Context Window, and FH is the Forecast Horizon.

\begin{table}[h]
    \centering
    % \caption{Hyperparameter Sweeps and Corresponding Test MSE for Top Model Versions}
    % \label{tab:hypersweeps}
    \begin{tabular}{l c}
        \hline
        \textbf{Model\_LR\_BS\_CW\_FH\_Epochs\_Optimizer} & \textbf{Test MSE} \\
        \hline
        ENV\_3e-3\_8\_24\_1\_200\_AdamW & \textbf{0.00010} \\
        ENV\_1.5e-3\_16\_24\_1\_500\_Adam & 0.00016 \\
        ENT\_3e-3\_16\_24\_1\_200\_AdamW & \textbf{0.00068} \\
        ENT\_5e-4\_32\_24\_1\_500\_Adam & 0.00075 \\
        ConvLSTM\_3e-3\_32\_24\_1\_200\_AdaGL & \textbf{0.00156} \\
        ConvLSTM\_1.5e-3\_16\_36\_1\_500\_Adam & 0.00192 \\
        MLP\_3e-3\_32\_24\_1\_200\_Adam & \textbf{0.06288} \\
        MLP\_1.5e-3\_16\_24\_1\_500\_AdamW & 0.06510 \\
        \hline
    \end{tabular}
    \caption{Hyperparameter sweeps and corresponding test MSE for the top-performing versions of each model. Hyperparameters include learning rate (LR), batch size (BS), context window (CW), forecast horizon (FH), epochs, and optimizer used.}
    \label{tab:hyperparam_mse}
\end{table}

These sweeps allowed us to identify the optimal settings for each architecture in terms of convergence speed and predictive accuracy.

\subsection*{7.3 Figures}
The following figures illustrate the training and validation performance of the models as well as key emissions visualizations.

\begin{figure}[h]
    \centering
    \includegraphics[width=0.5\textwidth]{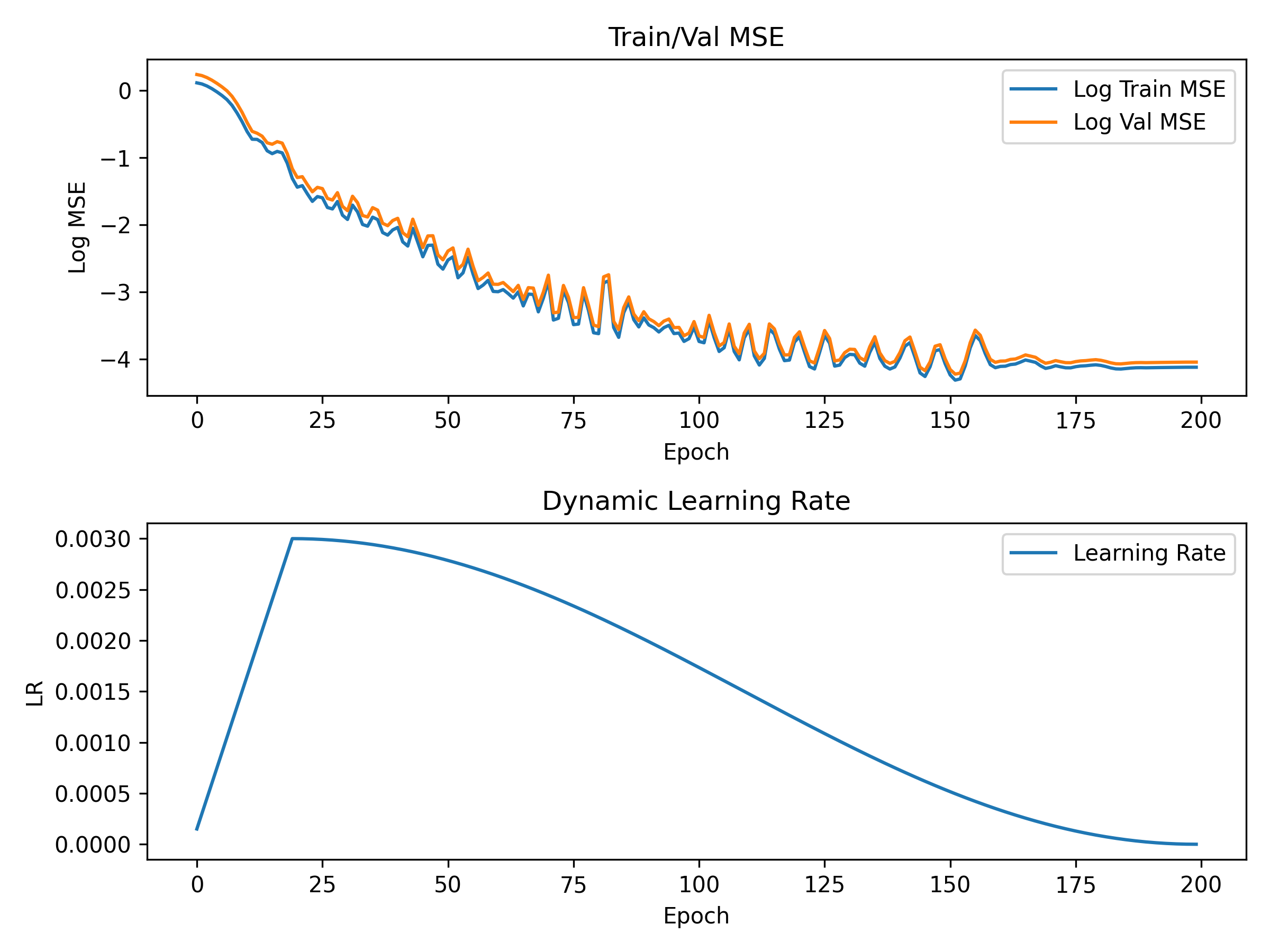}
    \caption{ENT Train and Validation MSE over Training Epochs}
    \label{fig:entv4_loss}
\end{figure}

\begin{figure}[h]
    \centering
    \includegraphics[width=0.5\textwidth]{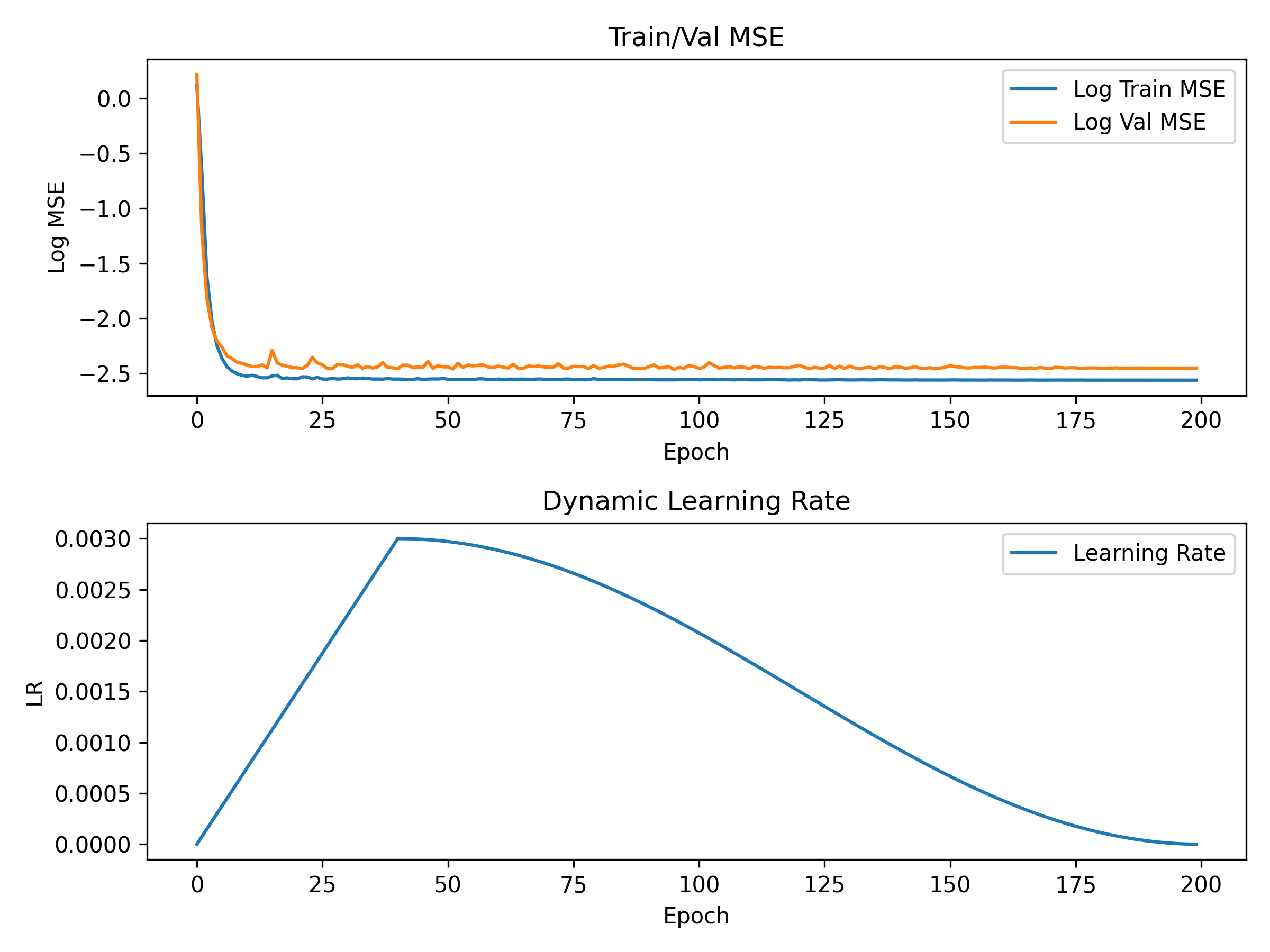}
    \caption{ConvLSTM Train and Validation MSE over Training Epochs}
    \label{fig:convlstm_loss}
\end{figure}

\begin{figure}[h]
    \centering
    \includegraphics[width=0.5\textwidth]{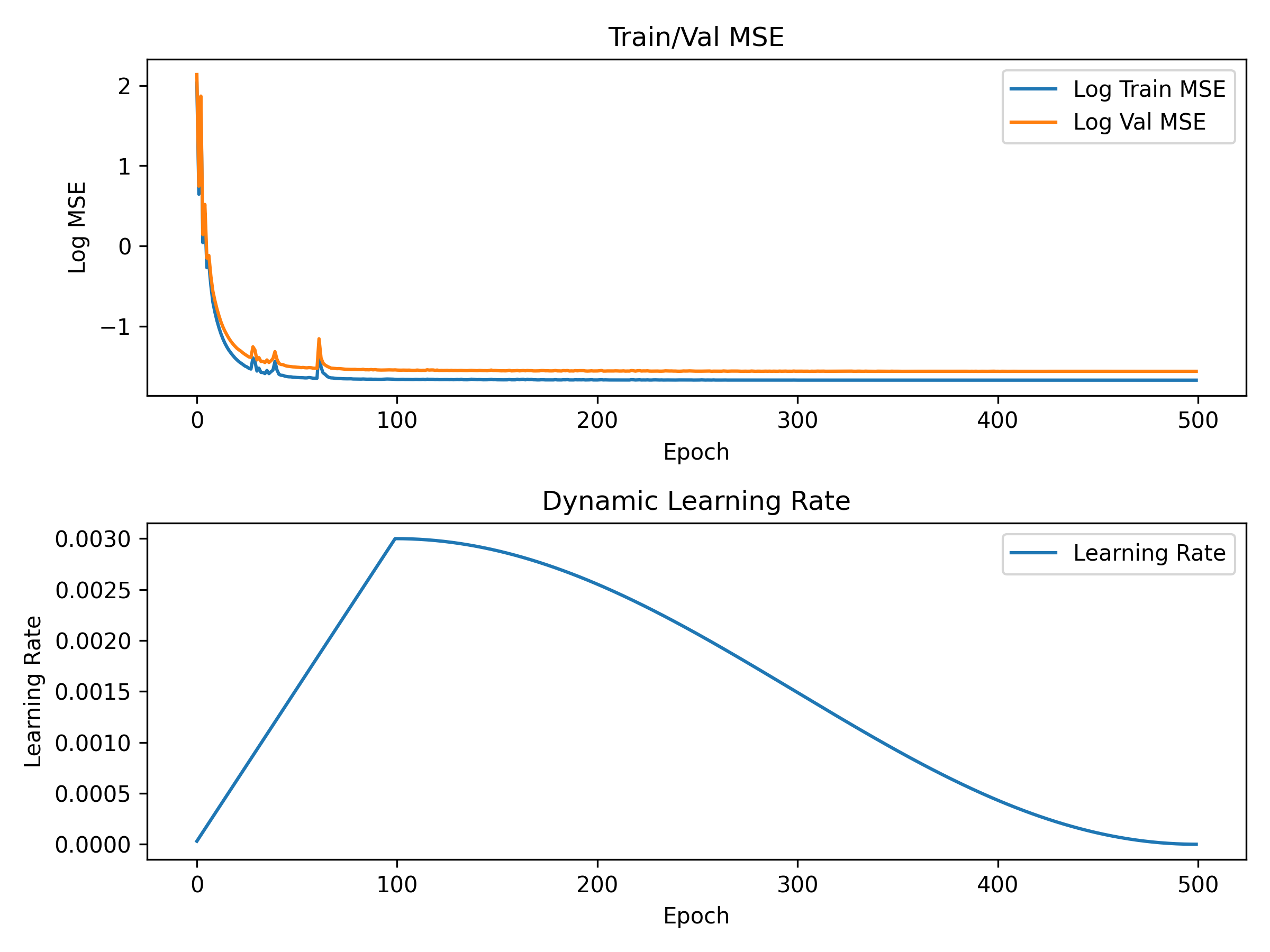}
    \caption{MLP Train and Validation MSE over Training Epochs}
    \label{fig:mlp_loss}
\end{figure}

\begin{figure}[h]
    \centering
    \includegraphics[width=0.5\textwidth]{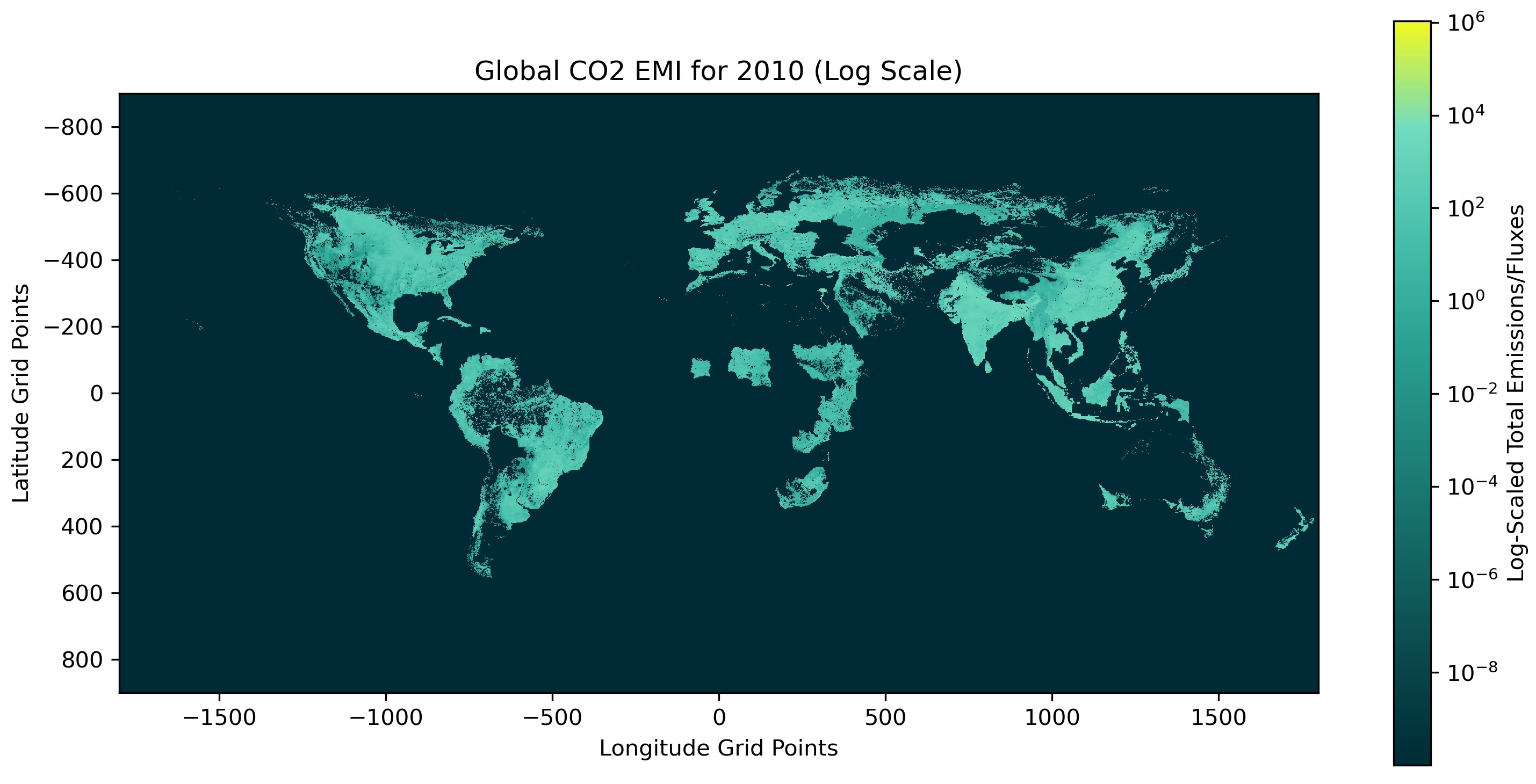}
    \caption{2010 Global CO$_2$ Emissions}
    \label{fig:co2}
\end{figure}

\begin{figure}[h]
    \centering
    \includegraphics[width=0.5\textwidth]{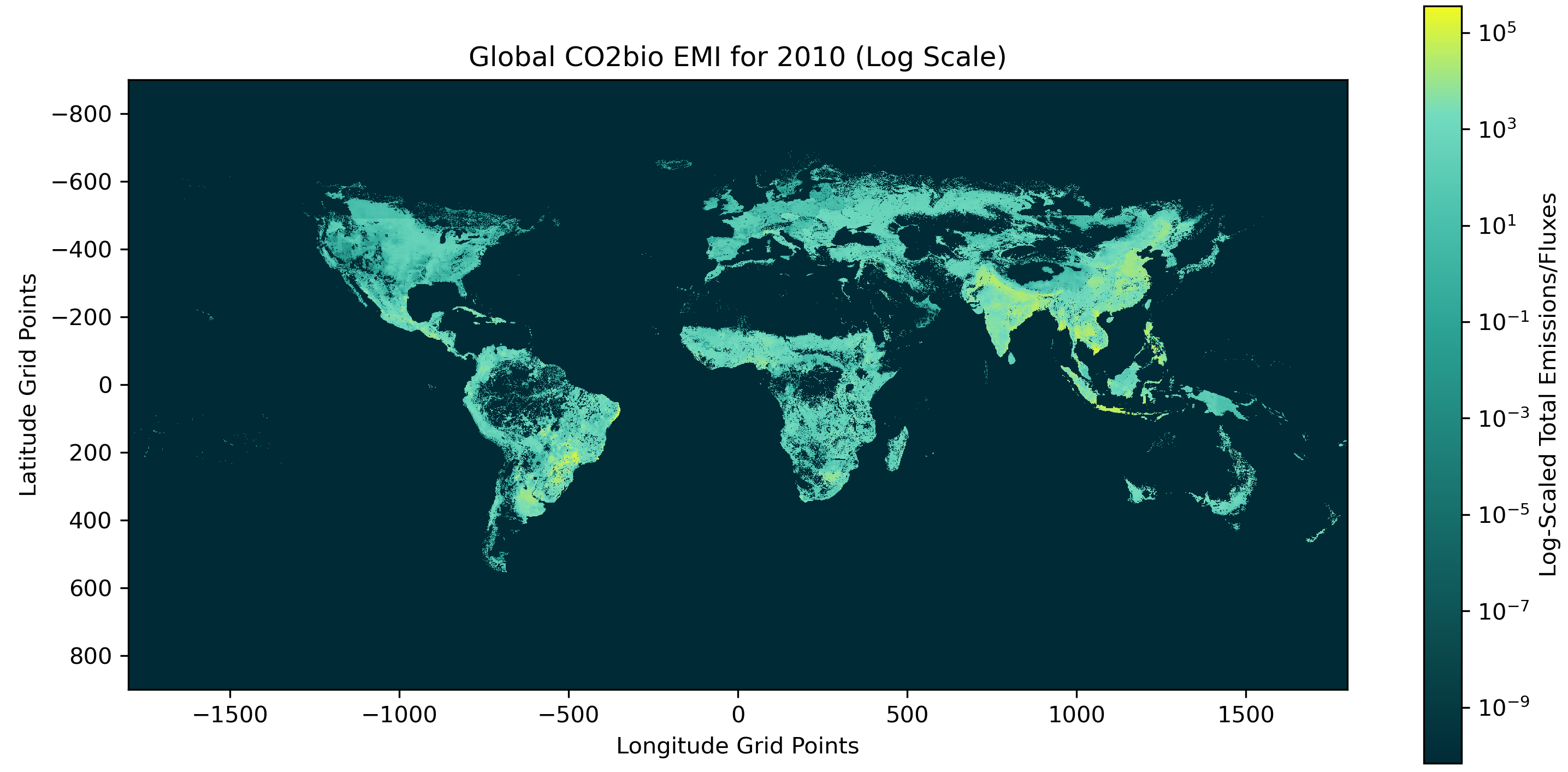}
    \caption{2010 Global CO$_2$Bio Emissions}
    \label{fig:co2bio}
\end{figure}

\begin{figure}[h]
    \centering
    \includegraphics[width=0.5\textwidth]{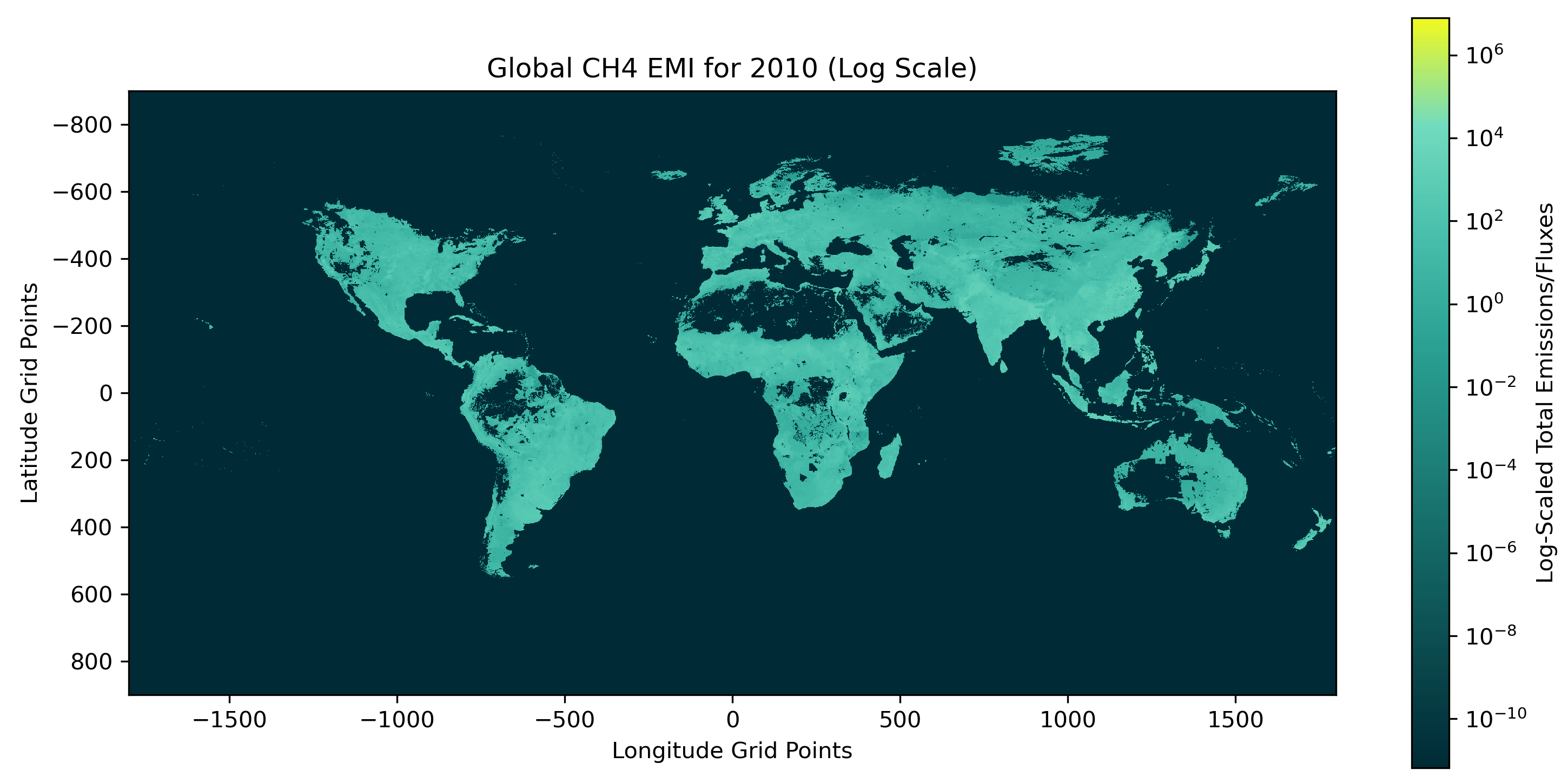}
    \caption{2010 Global CH$_4$ Emissions}
    \label{fig:ch4}
\end{figure}

\begin{figure}[h]
    \centering
    \includegraphics[width=0.5\textwidth]{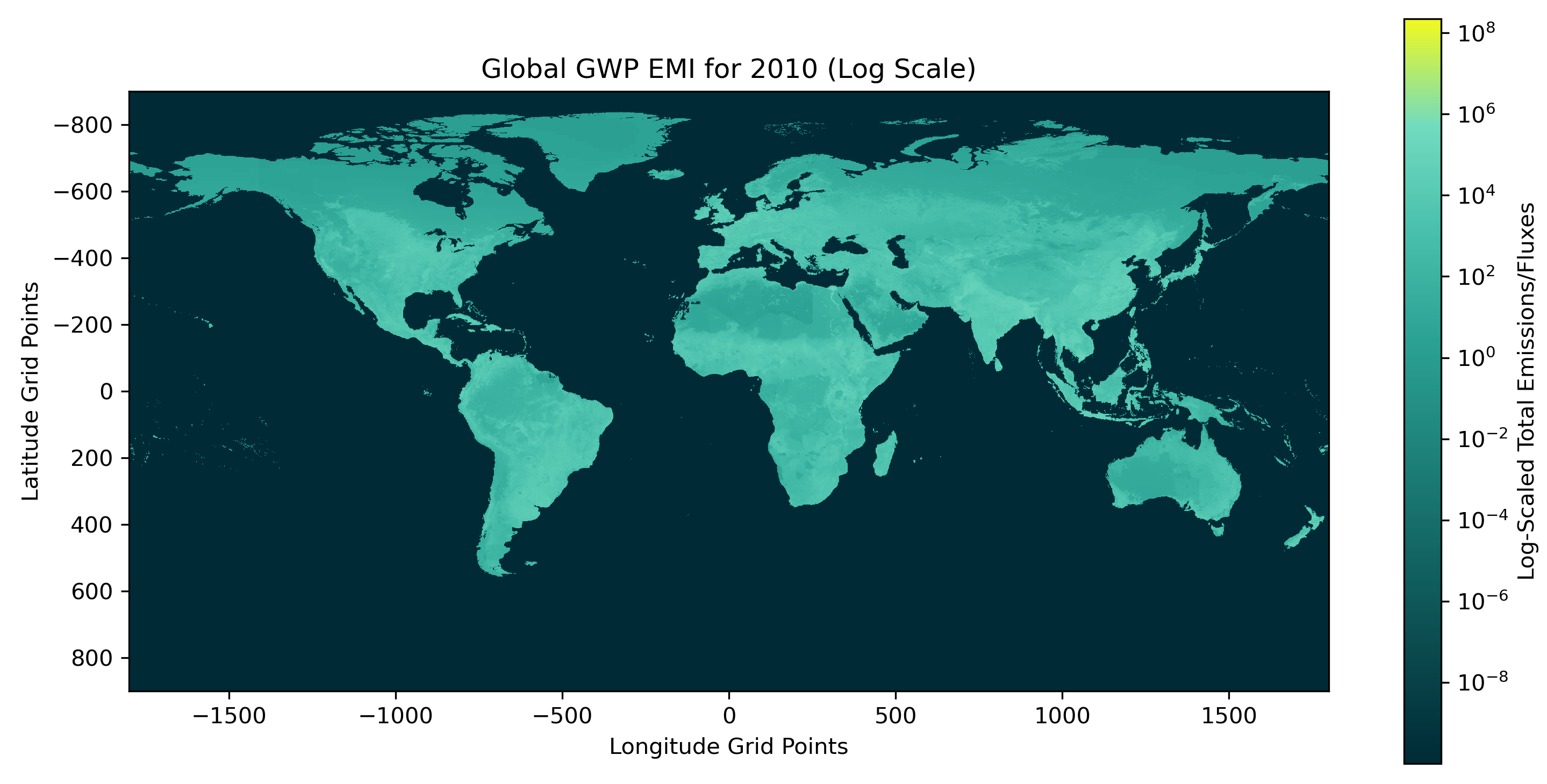}
    \caption{2010 Global GWP Emissions}
    \label{fig:gwp}
\end{figure}

%------------------------------------------------------

\end{document}